\documentclass[sigconf,screen]{acmart}

\usepackage{enumitem}  
\usepackage{stfloats} 
\usepackage{textcomp}
\usepackage{bbding}

\usepackage{MnSymbol}
\usepackage{graphicx}
\usepackage{subcaption}

\usepackage{booktabs}
\usepackage{lscape}        
\usepackage{longtable}
\usepackage{caption}
\usepackage{array}
\usepackage{multirow}
\usepackage[normalem]{ulem} 
\usepackage{makecell,pifont}
\usepackage{stfloats} 

\newcolumntype{C}[1]{>{\centering\arraybackslash}p{#1}} 

\usepackage[dvipsnames]{xcolor}
\usepackage{ifthen}
\newboolean{colorcomments}
\setboolean{colorcomments}{true} 

\newcommand{\bita}[1]{\ifthenelse{\boolean{colorcomments}}{\textcolor{Purple}{B: #1}}{#1}}
\newcommand{\john}[1]{\ifthenelse{\boolean{colorcomments}}{\textcolor{red}{John #1}}{#1}}

\AtBeginDocument{%
  }

\setcopyright{acmlicensed}
\copyrightyear{2026}
\acmYear{2026}
\acmDOI{XXXXXXX.XXXXXXX}
\acmConference[Conference acronym 'XX]{Make sure to enter the correct
  conference title from your rights confirmation email}{June 03--05,
  2018}{Woodstock, NY}

\acmISBN{978-1-4503-XXXX-X/2018/06}

\begin{document}

\title{IBAD: Interpretable Behavioral Anomaly Detection on Human Mobility Data}

\author{Bita Azarijoo}
\affiliation{%
  \institution{University of Southern California}
  \city{Los Angeles}
  \state{California}
  \country{USA}
}
\email{azarijoo@usc.edu}

\author{John Krumm}
\affiliation{%
  \institution{University of Southern California}
  \city{Los Angeles}
  \state{California}
  \country{USA}}
\email{jkrumm@usc.edu}

\author{Cyrus Shahabi}
\affiliation{%
  \institution{University of Southern California}
  \city{Los Angeles}
  \state{California}
  \country{USA}}
\email{shahabi@usc.edu}

\renewcommand{\shortauthors}{Azarijoo, Krumm, Shahabi}

\begin{abstract}
Human mobility appears highly diverse, yet much of a person's daily mobility can be explained by a small set of recurring behavioral templates, such as commuting, school-centered activities, caregiving, nightlife, or errand patterns. We present \texttt{IBAD} (\underline{I}nterpretable \underline{B}ehavioral \underline{A}nomaly \underline{D}etection), a framework that learns interpretable daily mobility templates and represents each individual as a distribution over mixtures of these templates. Rather than focusing on specific locations, IBAD characterizes activities that individuals perform across locations. This approach first discovers global behavioral templates using Latent Dirichlet Allocation (LDA), then employs a hierarchical self-supervised model to learn normal behavior of individuals from their soft behavioral templates. We also introduce a \emph{splicing benchmark} that creates controlled behavioral mismatches between an individual's historical profile and injected mobility patterns. Experiments on real-world and synthetic datasets show that daily behavior can be effectively decomposed into a small number of interpretable templates. Crucially, we show that the learned behavioral archetypes \emph{transfer} across distinct geographic and demographic contexts. Furthermore, IBAD maintains a robust competitive performance across all settings. For reproducibility purposes, the code is accessible at ~\href{https://github.com/USC-InfoLab/IBAD}{https://github.com/USC-InfoLab/IBAD}.
\end{abstract}



\maketitle

\section{Introduction}
Human mobility exhibits rich variability across individuals and days, yet much of this variation is structured rather than random. Daily activity patterns, such as home–work commutes, school-centered routines, caregiving trips, social outings, or multi-stop errands can be explained by a relatively small set of recurring behavioral templates. This motivates a new view of mobility anomaly detection: ordinary variation is not simply deviation from a fixed route (spatial) or schedule (temporal), but the natural behavioral diversity produced by recombining an individual's familiar templates. In contrast, a \emph{behavioral anomaly}, is a day whose mobility cannot plausibly be explained by the template mixture that characterizes that individual's historical behavior.

We propose \textbf{\texttt{IBAD}} (\underline{I}nterpretable \underline{B}ehavioral \underline{A}nomaly \underline{D}etection), a framework that learns interpretable daily mobility templates and represents each individual as a distribution over mixtures of these templates. This enables IBAD to distinguish natural variability, such as alternating between commuting, errands, and social-activity days, from unusual behavioral shifts, such as a routine commuter suddenly exhibiting an unexplained late-night multi-stop pattern or a school teacher showing a workday visit to an industrial zone\footnote{Walter White in Breaking Bad.}.  Instead of focusing on unusual locations or disruptions in temporal patterns exclusively, IBAD examines a population's general activities that occur at different locations. The approach begins by finding broadly descriptive, easily interpretable, daily behavioral templates with a topic modeling formulation over daily activity sequences, such as home-work-home. Each individual's daily behaviors can then be expressed as a weighted combination of the templates, enabling a structured decomposition of behavioral variability. To detect anomalies, IBAD further uses a hierarchical self-supervised framework to learn normal behavior with respect to both population-level templates and individual-specific template mixtures. We further introduce a splicing benchmark that evaluates whether a model can detect controlled behavioral mismatches between an individual's historical profile and an injected day or segment drawn from another behavioral archetype. Experiments on real-world and synthetic mobility datasets show that IBAD consistently outperforms state-of-the-art baselines while producing interpretable explanations of the templates underlying  injections.

IBAD's primary innovation is its focus on day-level behavioral templates rather than individual locations, timings, or transitions for anomaly detection on longitudinal human mobility data. While most mobility anomaly detection methods identify unusual places or movements, IBAD identifies days whose overall pattern of activities is inconsistent with an individual's historical behavioral repertoire. As a result, it is particularly well suited to detecting lifestyle shifts and behavioral changes that appear out of character for a person, even when no single  visit location is unusual. This matters because many applications care about changes in daily behavior rather than unusual locations: a person may visit only familiar places yet exhibit a day-long pattern inconsistent with their routines. Such changes are relevant to elder care~\cite{shahid2022anomalies,msaad2022routine,chifu2022routine}, cognitive decline monitoring~\cite{cejudo2025smart,civitarese2026xbcd,auker2021mobility}, psychiatric health~\cite{monk1990social,murray2009social}, child safety, fraud detection, and public safety~\cite{chen2026mtri,wen2025cobad}. Moreover, by
abstracting mobility into template mixtures rather than relying on handcrafted, prespecified dimensions, IBAD offers an extensible framework that can accommodate additional mobility components available in the underlying data.

\noindent We summarize our contributions as follows:
\begin{itemize}[leftmargin=*]
    \item \textbf{A novel behavioral paradigm.} We reframe mobility anomaly detection by shifting the focus from rigid spatial or temporal deviations to behavioral template mixtures. In this view, true anomalies are identified as days whose overall activity patterns cannot be explained by an individual's established routines.
    \item \textbf{Behavioral template discovery.} We show that  daily mobility of people decomposes into a small set of interpretable behavioral templates, and propose IBAD, a hierarchical self-supervised framework that scores each day by its deviation from the template mixture predicted from the individual's own history. 
    \item \textbf{Benchmarks for behavioral anomaly detection.} We release two
        resources: a semantically labeled Geolife benchmark, in which raw GPS stay-points are mapped to activity categories by combining OSM data with an LLM-as-a-Judge annotator; and a splicing protocol that injects
        within-person (\emph{intra}), cross-person (\emph{inter}), and combined behavioral mismatches, enabling evaluation without ground-truth anomaly labels.    
    \item \textbf{Robustness and transferability.} We show that the discovered templates learned independently on two datasets capture universal behavioral archetypes under a permutation test, indicating IBAD's robustness under data distribution shifts.
\end{itemize}

The rest of the paper proceeds as follows. Section~\ref{sec:related_works} surveys related work on
human mobility anomaly detection. Section~\ref{sec:template_discovery} explores decomposing daily mobility into interpretable behavioral templates. Section~\ref{sec:IBAD} details IBAD for day-level behavioral anomaly detection, and Section~\ref{sec:experiments} presents the experimental
evaluation. Section~\ref{sec:conclusion} concludes the paperwith a discussion of future directions.

\section{Related Work}
\label{sec:related_works}
Anomaly detection in spatiotemporal trajectories has long been an active research area in data mining. Depending on the nature of the trajectories under investigation, prior research can be broadly categorized into taxi trajectory anomaly detection and human mobility anomaly detection. This work focuses exclusively on the latter.

\subsection{Anomaly Detection in Taxi Trajectories}
Taxi trajectory anomaly detection, often used to identify fraudulent detours~\cite{chen2013iboat}, differs from general human mobility because taxis make variable, on-demand trips rather than following predictable routines. Early methods identified anomalous routes by comparing them to historical optima using partitioning, heuristics, or isolation algorithms~\cite{chen2013iboat,lee2008trajectory,zhang2011ibat,amiri2024urban}. Recently, deep learning has emerged to model complex spatiotemporal dependencies. For example, GM-VSAE~\cite{liu2020online} learns latent trajectory distributions via an RNN-based VAE, and ATROM~\cite{gao2023open} applies variational Bayesian and metric learning for open-world scenarios. Other approaches incorporate specific contexts: GETAD~\cite{mbuya2025graph} leverages road network topology and historical semantics, while FOTraj~\cite{shao2025towards} integrates LLMs with spatiotemporal encoders to manage real-time, noisy data streams.

\subsection{Anomaly Detection in Human Mobility}
Unlike taxi trajectory, anomaly detection in human mobility is a relatively new and still underexplored problem. In contrast to transportation networks, where optimal routes can often be defined, human mobility is inherently diverse, context-dependent, and difficult to characterize using a single notion of normal behavior. Existing research has largely focused on predefined forms of anomalous behavior, such as visits to uncommon locations, unusual timings at regular locations, deviations from historical mobility patterns, or engagement in specific activity types at unexpected times~\cite{zhang2024large,stanford2024numosim,amiri2024urban}.

\noindent Several recent methods have approached the problem from different perspectives. TOD4Traj~\cite{zhang2024transferable} employs contrastive learning to align spatiotemporal trajectory representations with semantic categories and regular temporal patterns, treating deviations from learned representations as anomalies. LMTAD~\cite{mbuya2024trajectory} formulates anomaly detection as a next-location prediction task using a language modeling framework, measuring the divergence between observed visits and model predictions. BeSTAD~\cite{xie2025bestad} identifies anomalies by combining multi-scale spatial semantics with variational clustering to detect deviations from a user's historically established routines across different time periods. However, its reliance on discrete behavioral clusters and aggregated temporal segments limits its ability to represent continuous, overlapping, and multi-intent daily activities that commonly arise in real-world mobility patterns. ICAD~\cite{azarijoo2025icad} detects visit-level and agent-level anomalies by scoring visits relative to highly probable normal routines. While effective at identifying spatial and fine-grained temporal deviations, it largely assumes that anomalous behavior manifests through deviations in only location or time, potentially oversentive to natural behavioral variations. Taking a different approach, CoBAD~\cite{wen2025cobad} shifts the focus from individual trajectories entirely to collective human mobility. It models the spatiotemporal dependencies and interactions across multiple individuals, employing a two-stage attention mechanism on a co-occurrence event graph to identify collective anomalies, such as unexpected co-occurrences or abnormal absences. Finally, TrajGenAgent~\cite{li2026trajgenagent} is a zero-shot agentic framework originally designed for mobility generation. However, to evaluate, it uses ICAD and BeSTAD for visit-level and agent-level anomalies, respectively.

\subsection{Topic Modeling for Mobility}
A separate line of work adapts topic models to activity-travel data to recover
interpretable structure from unlabeled mobility records. Building on LDA~\cite{blei2003latent}, Zhao et
al.~\cite{zhao2020spatiotemporal} treat each individual's activity history as a
document and each activity episode as a word, recovering latent activity types
such as home and work as topics over space and time, an idea later extended with
land-use context~\cite{smartcard_landuse} and used for travel-behavior anomaly
detection~\cite{zhao2018lpr_anomaly}. However, this lineage models the
\emph{individual} as the unit of analysis at the level of trips or episodes,
yielding a static mixture over activity types. In contrast, IBAD takes
each agent-day as the unit of analysis, modeling how these routines evolve over time to flag days inconsistent
with an individual's history.

\section{Behavioral Template Discovery}
\label{sec:template_discovery}

We speculate that complex human mobility patterns follow a relatively small number of underlying daily templates. If this holds true, we can leverage these underlying structures to study an underexplored yet important category of behavioral anomalies in a new abstraction level. In this section, we first formalize the notion of a behavioral template and develop a discovery procedure that learns these archetypes in a fully unsupervised manner from
agents' daily activity timelines, yielding a compact mixture of interpretable templates that explain a significant portion of an agent's daily behavior. To this end, we demonstrate the discovery process on real-world travel survey data, verifying the emergence of semantically meaningful templates and establishing the foundational representations that support our anomaly detection framework detailed in Section~\ref{sec:IBAD}.

\subsection{Problem Formulation}
\label{sec:template_problem_formulation}
\noindent\paragraph{Preliminaries.} Let $\mathcal{D} = \{\mathbf{x}_d\}_{d=1}^{D}$ denote a
dataset of $D$ \emph{agent-days} pooled across all agents, where each $\mathbf{x}_d$ is
the activity \emph{timeline} of one agent on a single day, encoded as a sequence over
$S$ evenly spaced discrete time slots:
\begin{equation}
  \mathbf{x}_d = \bigl(x_d^{(1)}, \dots, x_d^{(S)}\bigr),
  \qquad x_d^{(s)} \in \mathcal{A},
\end{equation}
where $x_d^{(s)}$ is the activity occupying slot $s$ and $\mathcal{A}$ is a finite set of
activity types (e.g., \texttt{Home}, \texttt{Work}, \texttt{Education}). We refer to each $\mathbf{x}_d$ as an
\emph{agent-day}. Template discovery pools all agent-days, so we do not include the agent's identity in the notation.
 
To capture the structural regularities within these sequences, we model them using \emph{behavioral templates}, which are prototypical daily patterns shared across many agent-days, such as a student spending most of their day time in school. Because a single day rarely conforms to one prototype in isolation (a workday may include an afternoon visit to gym), we do not rigidly assign each day to a single template but rather a \emph{soft membership} over templates.

\smallskip
\noindent\paragraph{Problem (Behavioral Template Discovery).}
\emph{Given} an unlabeled dataset $\mathcal{D}$ of agent-days and a number of
templates $K$, we aim to \emph{learn} two things: (i)~a set of $K$ behavioral
templates $\mathcal{T} = \{\mathcal{T}_k\}_{k=1}^{K}$, and (ii)~for each
agent-day $d$, a soft membership vector:
\begin{equation}
  \theta_{d,k} \ge 0, \quad \sum_{k=1}^{K} \theta_{d,k} = 1,
\end{equation}
where $\theta_{d,k}$ measures how strongly template $k$ explains agent-day $d$. The templates $\mathcal{T}$ form a shared basis of
daily archetypes, and each  $\boldsymbol{\theta}_d$ then expresses one
agent-day as a mixture over that basis. The task of discovering the templates is fully unsupervised, and each $\boldsymbol{\theta}_d$
serves as a compact, interpretable summary of a single day's behavior, serving as ground truth for the IBAD anomaly detector. 
For instance,
with $K=3$ templates capturing, say, a \emph{work} day
$\mathcal{T}_1$, a \emph{leisure} day $\mathcal{T}_2$,
and a \emph{stay-at-home} day $\mathcal{T}_3$, an agent-day that is
mostly a standard working day with an afternoon shopping trip
might be summarized as $\boldsymbol{\theta}_d = (0.7,\, 0.25,\, 0.05)$,
indicating that template~1 dominates while template~2 contributes a
smaller share and template~3 is almost absent. Figure~\ref{fig:template_extraction} illusterates this.

\subsection{Behavioral Template Extraction via LDA}
\label{sec:template_extraction}
We cast behavioral template extraction as a \emph{topic modeling} problem over daily activity timelines~\cite{blei2003latent}. Originally developed for text, topic modeling represents documents as latent topics, and each topic is characterized by a distribution over words. We adapt this framework to human mobility by treating behavioral templates as latent topics and daily activity timelines as documents. Under this formulation, model fitting simultaneously discovers latent behavioral templates and estimates the extent to which each template is expressed within an individual day.

\paragraph{Activity timeline construction from mobility records.}

To obtain a uniform representation across individuals and days, we first transform raw mobility records into fixed-length daily activity timelines. We assume that, for each agent, the input is given as a list of time slots together with the activity occurred during each slot. Following experimental setup of Appendix~\ref{sec:lda_templates}, each day is divided into $S$ fixed daily time slots, and every slot is assigned the activity that occupies it. All slots are initially assigned a default \texttt{Home} activity. This procedure produces a complete daily activity sequence in which every time slot is associated with exactly one activity. For any reported non-\texttt{Home} activity, we overwrite the corresponding slots with that activity. This procedure produces a complete daily activity sequence in which every time slot is associated with exactly one activity.

\paragraph{Time-aware tokenization.}
To preserve both activity type and temporal context, we represent each time slot by a
time-stamped activity token $w=(s,a)$, where $s$ is the time slot and $a \in \mathcal{A}$
is the activity category. This gives the vocabulary
\begin{equation}
\mathcal{V}=\{1,\ldots,S\}\times\mathcal{A},
\qquad
|\mathcal{V}|=S\cdot M,
\end{equation}
where $M=|\mathcal{A}|$ is the number of activity categories. By folding the time index
into the token itself, we let the same activity at different times of day act as distinct
tokens. Each agent-day is therefore a sequence of $S$ tokens, one per time slot.

\paragraph{LDA behavioral template modeling.}
Given this representation, we fit the corpus of agent-days with Latent Dirichlet Allocation~\cite{blei2003latent}. Each template $k$ is a distribution $\boldsymbol{\phi}_k$ over the
vocabulary $\mathcal{V}$, where $\phi_{k,(s,a)}$ is the probability of activity $a$ at time
slot $s$; a template thus acts as a soft prototype of a full day. Each agent-day $d$ in turn
receives a distribution $\boldsymbol{\theta}_d$ over the $K$ templates, where $\theta_{d,k}$
is the proportion of the day attributed to template $k$. Both are given Dirichlet priors,
$\boldsymbol{\phi}_k \sim \mathrm{Dir}(\beta)$ and
$\boldsymbol{\theta}_d \sim \mathrm{Dir}(\alpha)$, which favor sparse templates and sparse
per-day mixtures.

The model generates a day, one slot at a time: for each slot $n$ it draws a template
$z_{d,n} \sim \mathrm{Cat}(\boldsymbol{\theta}_d)$ and then emits the token
$w_{d,n} \sim \mathrm{Cat}(\boldsymbol{\phi}_{z_{d,n}})$. Although LDA treats each document as
an unordered bag of words, temporal information is preserved as every token carries its
own time slot, the same activity at different hours is a distinct word, so the timing of
each activity is preserved within the vocabulary itself.

\begin{figure}[t]
    \centering
    \includegraphics[width=\linewidth]{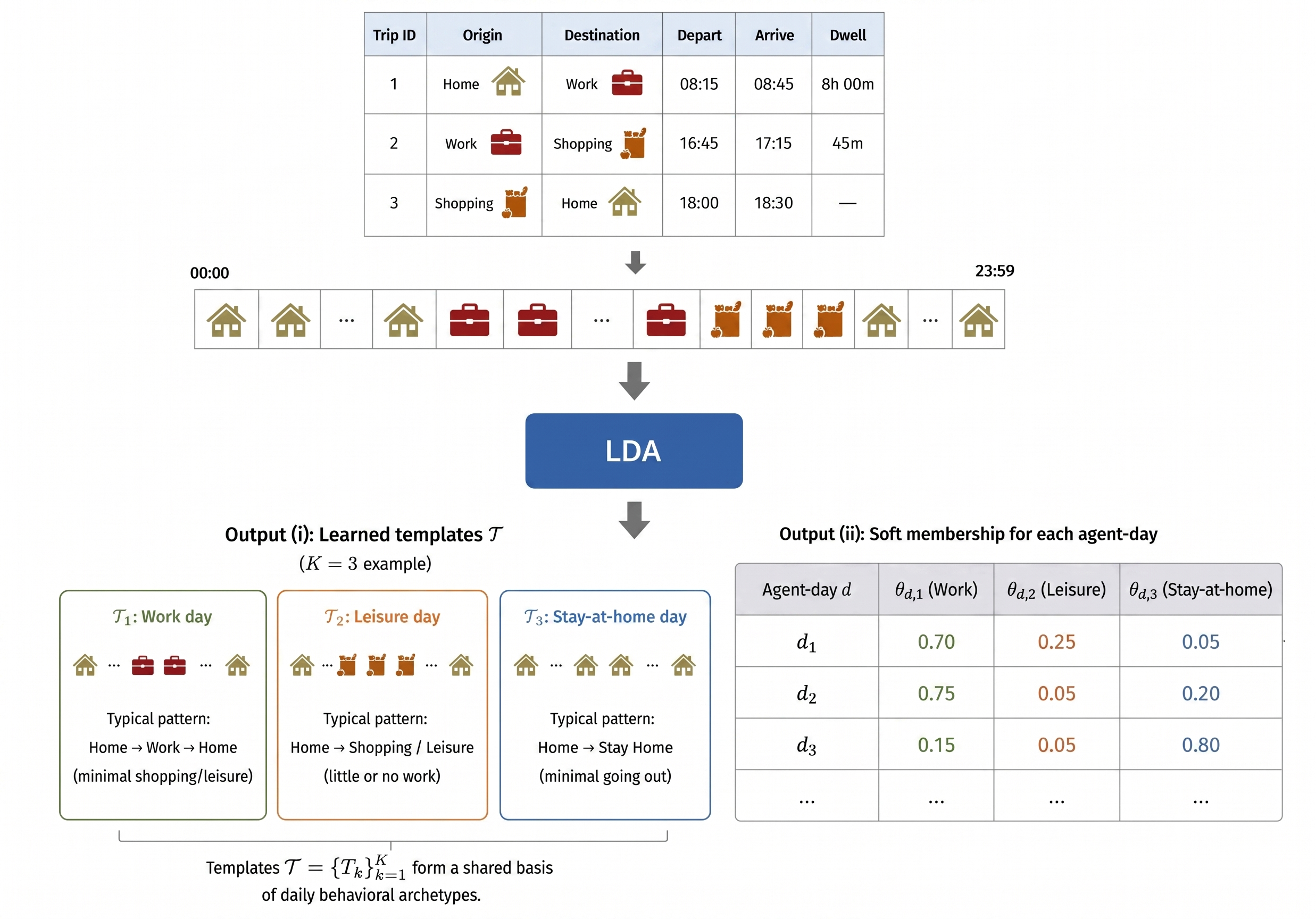}
    \caption{Illustration of behavioral template discovery.} 
    \label{fig:template_extraction} 
\end{figure}

\paragraph{Inference and representation.}
Fitting the model means estimating the template distributions $\boldsymbol{\phi}_k$ and the
per-day mixtures $\boldsymbol{\theta}_d$ from the observed activity tokens. Computing them
exactly requires the posterior distribution over all templates and mixtures given the
corpus, which has no closed form and is intractable at our scale. We therefore approximate
it with Online Variational Bayes~\cite{hoffman2010online}, which processes the data in
mini-batches and scales comfortably to large mobility datasets. Fitting returns the learned
templates $\{\boldsymbol{\phi}_k\}_{k=1}^{K}$ and an estimated mixture
$\hat{\boldsymbol{\theta}}_d$ for every day of each agent, with the number of templates $K$ prespecified, as visualized in Figure~\ref{fig:template_extraction}.

Each template $\boldsymbol{\phi}_k$ can be reshaped into an $M \times S$ time-by-activity
matrix, giving an interpretable behavioral archetypes. The estimated mixtures $\hat{\boldsymbol{\theta}}_d$ act as compact behavioral signatures of an agent's day and serve as the ground truth for IBAD in Section~\ref{sec:IBAD}.
.
\subsection{Empirical Observations}
\label{subsec:empirical_observations_templates_NHTS}
In this section, we evaluate our template discovery framework using the 2017 National Household Travel Survey (NHTS) to identify latent behavioral structures. The NHTS dataset is highly suited for this task: it captures a large, demographically diverse U.S. population and provides a rich taxonomy of trip purposes. This allows us to test whether complex daily routines naturally decompose into a compact set of interpretable templates that define normal behavior. Note that due to privacy constraints, the dataset does not include precise spatial coordinates for trips. Comprehensive details regarding the NHTS dataset and our experimental setup are provided in Sections~\ref{sec:NHTS_dataset} and~\ref{sec:template_discovery}, respectively.

Figure~\ref{fig:nhts_templates} shows $K{=}6$ templates recovered from NHTS~2017. Each subplot shows the probability distribution of an activity occurring at a particular time slot. For brevity, we only show the top three dominant activities for each subplot. For instance, Template~4 captures full-time
9-to-5 worker behavior: \texttt{Home} declines around 08:00, \texttt{Work} dominates through midday, and \texttt{Home} recovers
near 17:00. Template~5 captures a student day, with an analogous \texttt{Home}-to-\texttt{Education} shift
aligned to school hours. Template~0 captures an evening or late-work pattern that peaks near
18:00 and probably belongs to shift workers. Template~1 groups midday \texttt{Exercise} and \texttt{Recreation} trips, consistent with retired adults. Template~2 is dominated by afternoon \texttt{Recreation} and \texttt{Social} activity,
characteristic of weekend days. Template~3 corresponds to a stay-at-home day. The templates
are also temporally well separated, in that their non-\texttt{Home} peaks occupy distinct windows of
the day, and they recur across tens of thousands of person-days from heterogeneous
households. Together these observations confirm that a small, shared set of templates
explains the bulk of normal daily mobility.

Crucially, this shared vocabulary does not confine each individual to a 
single template. A typical agent traverses a \emph{mixture} of templates 
over time: a full-time worker may spend weekdays in Topic~4, occasional 
evenings in Topic~0, and weekends in Topic~2 or Topic~3. What 
characterizes the agent is not any single template but their distribution 
over templates. This induced a daily behavioral profile that remains consistent as long as the underlying routine is preserved.

This view motivates our target task: \emph{day-level behavioral anomaly 
detection}, in which individual days are flagged when they deviate substantially from 
their regular mixture of templates. In the next section, we explain the architecture of IBAD, which learns each agent's behavioral characteristic during training and and scores each day by its consistency with that mixture during inference. For fairness, IBAD does not use location coordinates, since they were unavailable in the travel survey due to privacy constraints.

\section{IBAD Architecture}
\label{sec:IBAD}
This section presents IBAD, an architecture for detecting day-level 
behavioral deviations relative to an agent's established mixture of 
expected daily routines. We first formalize behavioral anomaly detection 
on \emph{longitudinal} human mobility data in Section~\ref{sec:IBAD_problem_formulation}. 
We then describe IBAD's predictive model, which estimates the 
template mixture of a target day conditioned on the agent's prior 
history. Finally, we discuss the anomaly scoring method, which measures relative divergence between observed behavioral mixtures and the model's prediction.

\begin{figure*}[t]
    \centering
    \includegraphics[width=\linewidth]{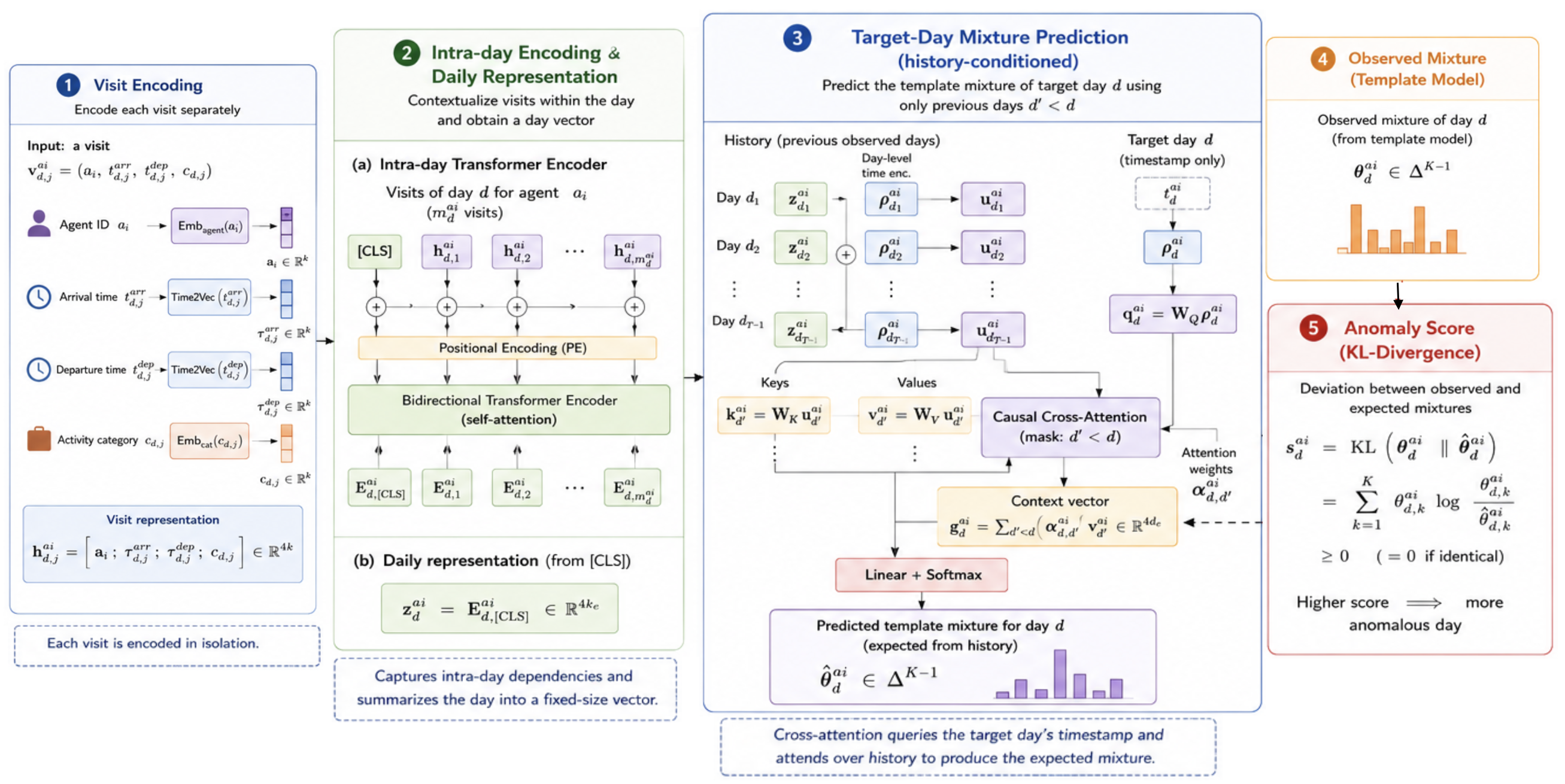}
    \caption{Overview of IBAD architecture components. (1) IBAD embeds each visit component in isolation. (2) Each visit of the day is encoded through a bidirectional transformer encoder, and a [CLS] token is added to obtain a daily summary representation of the day. (3) A history-conditioned causal encoder predicts the template mixture for a target day, based on  representations of previous days. (4) and (5) Finally, IBAD computes an anomaly score by calculating the KL-divergence between the observed mixture (4) and the estimated template distribution, using this score to optimize the network.} 
    \label{fig:IBAD} 
\end{figure*}

\subsection{Problem Formulation}
\label{sec:IBAD_problem_formulation}

We consider a longitudinal mobility dataset of $N$ agents $\{a_i\}_{i=1}^{N}$ observed
over a fixed time horizon. For agent $a_i$ ($i = 1,\dots,N$) on day $d$, let:
\[
\ell^{a_i}_d = \bigl[\, v^{a_i}_{d,1},\, v^{a_i}_{d,2},\, \dots,\, v^{a_i}_{d,m^{a_i}_d}\,\bigr]
\]
denote the ordered sequence of visits made on that day, where $m^{a_i}_d$ is the number
of visits and the $j$-th visit ($j = 1,\dots,m^{a_i}_d$):
\[
v^{a_i}_{d,j} = \bigl(t^{\text{arr}}_{d,j},\; t^{\text{dep}}_{d,j},\; c_{d,j}\bigr)
\]
records arrival time, departure time, and activity category $c_{d,j} \in \mathcal{A}$. This corresponds to realistic travel survey data, synthetic trajectories, or raw GPS data that has been processed into trips. The agent's full history up to day $d$ is the 
sequence of daily visits $\mathcal{H}^{a_i}_{<d} = \bigl(\ell^{a_i}_1, 
\ell^{a_i}_2, \dots, \ell^{a_i}_{d-1}\bigr)$. Following Section~\ref{sec:template_discovery}, each day $d$ is also associated 
with a soft template assignment $\theta^{a_i}_d \in \Delta^{K-1}$, the 
posterior mixture over the $K$ behavioral templates induced by $\ell^{a_i}_d$ 
under the template model. The objective of day-level behavioral anomaly 
detection is to assign each day $d$ a score $s^{a_i}_d \in \mathbb{R}$ 
that quantifies how inconsistent $\theta^{a_i}_d$ is with the mixture 
expected from $\mathcal{H}^{a_i}_{<d}$, i.e. the agent's history. Figure~\ref{fig:IBAD} illustrates the training pipeline.

\subsection{IBAD Training Pipeline}
\label{sec:ibad-training}
\subsubsection{Visit Embedding}
\label{sec:ibad_visit_embedding}

We first describe how each individual's visit is mapped into a latent space. Consider a single visit $v^{a_i}_{d,j} = (a_i,\, t^{\text{arr}}_{d,j},\, 
t^{\text{dep}}_{d,j},\, c_{d,j})$. Each of its four components is 
encoded independently into $\mathbb{R}^k$.

\paragraph{Agent identity.}
The agent identifier $a_i$ is mapped to a learnable embedding vector
\begin{equation}
   \mathbf{a}_i = \mathrm{Emb}_{\text{agent}}(a_i) \in \mathbb{R}^{k},
   \label{eq:agent-emb}
\end{equation}
shared across all visits of agent $a_i$. The agent embedding lets 
downstream layers condition on agent-level behavioral priors.

\paragraph{Temporal encoding.}
The arrival and departure timestamps are encoded with 
Time2Vec~\cite{kazemi2019time2vec}, which captures both periodic and 
non-periodic temporal patterns:
\begin{equation}
   \boldsymbol{\tau}^{\text{arr}}_{d,j} = \mathrm{Time2Vec}\bigl(t^{\text{arr}}_{d,j}\bigr), 
   \qquad
   \boldsymbol{\tau}^{\text{dep}}_{d,j} = \mathrm{Time2Vec}\bigl(t^{\text{dep}}_{d,j}\bigr),
   \label{eq:time2vec}
\end{equation}
with $\boldsymbol{\tau}^{\text{arr}}_{d,j},\,\boldsymbol{\tau}^{\text{dep}}_{d,j} \in \mathbb{R}^{k}$.

\paragraph{Activity category.}
The activity category $c_{d,j}$ is mapped to a learnable embedding 
vector
\begin{equation}
   \mathbf{c}_{d,j} = \mathrm{Emb}_{\text{cat}}(c_{d,j}) \in \mathbb{R}^{k},
   \label{eq:cat-emb}
\end{equation}
parameterized independently from the agent embedding.

\paragraph{Visit-level representation.}
The four component encodings are concatenated along the feature 
dimension to form the visit's representation:
\begin{equation}
   \mathbf{h}^{a_i}_{d,j} =
   \bigl[\,\mathbf{a}_i \,;\, \boldsymbol{\tau}^{\text{arr}}_{d,j} \,;\, 
   \boldsymbol{\tau}^{\text{dep}}_{d,j} \,;\, \mathbf{c}_{d,j}\,\bigr]
   \in \mathbb{R}^{4k},
   \label{eq:visit-emb}
\end{equation}
where $[\,\cdot\,;\,\cdot\,]$ denotes concatenation along the feature 
dimension.

\subsubsection{Daily Behavior Representation} 
To obtain a rich contextual daily behavioral representation of an agent, we employ a hierarchical approach. First, we contextualize the visits of each day
with a bidirectional Transformer encoder, so that every visit is informed by the others
made on the same day, and pool the resulting sequence into a single fixed-size day vector
$\mathbf{z}^{a_i}_d$ via a learnable \texttt{[CLS]} token.

\paragraph{Intra-day visit encoding.}
The per-visit embeddings of Section~\ref{sec:ibad_visit_embedding} are computed in
isolation and carry no information about the other visits of the day. To capture the
semantic and temporal dependencies \emph{among} a day's visits, we contextualize them with
a bidirectional Transformer encoder. For agent $a_i$ on day $d$, we prepend a learnable
[CLS] token $\mathbf{h}_{\text{[CLS]}} \in \mathbb{R}^{4k_e}$ and stack the day's visit
embeddings into
\begin{equation}
   \mathbf{H}^{a_i}_d =
   \bigl[\, \mathbf{h}_{\text{[CLS]}};\, \mathbf{h}^{a_i}_{d,1};\, \dots;\,
   \mathbf{h}^{a_i}_{d,m^{a_i}_d} \,\bigr]
   \in \mathbb{R}^{(m^{a_i}_d + 1) \times 4k_e},
\end{equation}
where each visit row is the representation from Eq.~\eqref{eq:visit-emb}. Self-attention is
permutation-invariant, so we add positional encodings $\mathrm{PE}(\cdot)$ to retain the
order of visits within the day~\cite{vaswani2017attention}, and we apply the encoder
bidirectionally so that every visit can attend to both earlier and later visits of the same
day:
\begin{equation}
   \mathbf{E}^{a_i}_d =
   \mathrm{TransformerEncoder}\bigl(\mathbf{H}^{a_i}_d + \mathrm{PE}(\mathbf{H}^{a_i}_d)\bigr)
   \in \mathbb{R}^{(m^{a_i}_d + 1) \times 4k_e}.
\end{equation}
Each row of $\mathbf{E}^{a_i}_d$ is a contextualized encoding of the corresponding visits in a day.

\paragraph{Daily representation via the \texttt{[CLS]} token.}
We summarize the day into a single vector by reading out the contextualized [CLS] row,
\begin{equation}
   \mathbf{z}^{a_i}_d = \mathbf{E}^{a_i}_{d,\text{[CLS]}} \in \mathbb{R}^{4k_e}.
\end{equation}
Through self-attention, $\mathbf{z}^{a_i}_d$ aggregates information from all visits of day
$d$ into a fixed-size behavioral representation, independent of daily visit count
$m^{a_i}_d$. Concretely, the attention weights assigned to the \texttt{[CLS]} token determine how much each
visit contributes to $\mathbf{z}^{a_i}_d$, allowing the representation to emphasize the
visits that most explains dominant daily archetypes of a person while down-weighting less important visits. For instance, consider a student whose day primarily consists of attending classes, but had a brief visit to a grocery store. In this setting, the self-attention mechanism will assign higher weights to prolonged school visits, as they strongly characterize the \texttt{Education} template. Consequently, $\mathbf{z}^{a_i}_d$ robustly captures the person's behavioral archetypes without being skewed by minor variations. 

\subsubsection{Target-Day Mixture Prediction}
\label{sec:target_day_pred}
The day-level vectors $\mathbf{z}^{a_i}_d$ summarize each day in
isolation, with no knowledge of the days around it. IBAD predicts the behavioral template
mixture of a \emph{target} day from the agent's earlier days, while conditioning explicitly
on \emph{when} the target day occurs. We implement this with a causal cross-attention layer:
the query is derived from the target day's timestamp, and the keys and values are the
representations of the agent's previous days.

\paragraph{Day-level temporal encoding.}
Each day carries a calendar timestamp $t^{a_i}_d$, which we encode with
Time2Vec~\cite{kazemi2019time2vec},
\begin{equation}
   \boldsymbol{\rho}^{a_i}_d = \mathrm{Time2Vec}\bigl(t^{a_i}_d\bigr) \in \mathbb{R}^{4k_e},
\end{equation}
where $\boldsymbol{\rho}^{a_i}_d$ is the day-level temporal encoding, distinct from the
intra-visit encodings $\boldsymbol{\tau}^{\text{arr}},\boldsymbol{\tau}^{\text{dep}}$ of
Section~\ref{sec:ibad_visit_embedding}.

\paragraph{History-conditioned cross-attention.}
To predict the mixture of target day $d$, we form a query from \emph{only} the target day's
timestamp and attend over the agent's earlier days $d' < d$, each represented by its
behavioral vector combined with its own temporal encoding:
\begin{equation}
\begin{aligned}
   \mathbf{u}^{a_i}_{d'} &= \mathbf{z}^{a_i}_{d'} + \boldsymbol{\rho}^{a_i}_{d'}, \\
   \mathbf{q}^{a_i}_d    &= \mathbf{W}_Q\,\boldsymbol{\rho}^{a_i}_d, \\
   \mathbf{k}^{a_i}_{d'} &= \mathbf{W}_K\,\mathbf{u}^{a_i}_{d'}, \\
   \mathbf{v}^{a_i}_{d'} &= \mathbf{W}_V\,\mathbf{u}^{a_i}_{d'}.
\end{aligned}
\end{equation}
A causal mask restricts the attention to $d' < d$, so the prediction depends strictly on the
history $\mathcal{H}^{a_i}_{<d}$ and never on the target day's own behavior. The attended
context is
\begin{equation}
   \mathbf{g}^{a_i}_d
   = \sum_{d' < d} \alpha^{a_i}_{d,d'}\,\mathbf{v}^{a_i}_{d'},
   \qquad
   \alpha^{a_i}_{d,d'}
   = \mathrm{softmax}_{d' < d}\!\left(
        \frac{\mathbf{q}^{a_i}_d \cdot \mathbf{k}^{a_i}_{d'}}{\sqrt{4d_e}}
     \right),
\end{equation}
where $\mathbf{g}^{a_i}_d \in \mathbb{R}^{4d_e}$ captures the agent's expected behavior at the
target day's position in time. In practice we stack several such layers into a causal
Transformer decoder, but a single head is shown in Figure~\ref{fig:IBAD} for clarity.

Conditioning the query on the target timestamp is deliberate, because mobility is recorded
\emph{irregularly}: an agent is not observed on every day of the horizon, so the gap between
consecutive observed days varies. Querying by the target day's time lets the model place its
prediction at the correct point on the timeline, weigh recent days against distant ones, and
remain robust to mild, time-localized behavioral drift, for example a relocation after which
the agent resumes a similar routine in a new place. Thus IBAD seamlessly works when some days in the agent's recorded history are missing.

\paragraph{Mixture Prediction.}
A linear layer with a softmax maps the attended context to a distribution over the $K$
behavioral templates,
\begin{equation}
   \hat{\boldsymbol{\theta}}^{a_i}_d
   = \mathrm{softmax}\bigl(\mathbf{W}_o\,\mathbf{g}^{a_i}_d + \mathbf{b}\bigr)
   \in \mathbb{R}^{K},
   \qquad
   \hat{\theta}^{a_i}_{d,k} \ge 0, \quad \sum_{k=1}^{K}\hat{\theta}^{a_i}_{d,k} = 1,
\end{equation}
so $\hat{\boldsymbol{\theta}}^{a_i}_d$ is IBAD's \emph{predicted} template mixture for target
day $d$: the blend of behavioral routines that the agent's history leads us to expect before that
day is observed.

\subsubsection{Training Objective}
\label{sec:ibad_training_objective}
We supervise the predicted mixture $\hat{\boldsymbol{\theta}}^{a_i}_d$ against the observed
mixture $\boldsymbol{\theta}^{a_i}_d$ that the template model assigns to day $d$
(Section~\ref{sec:template_discovery}). Since both are distributions over the $K$ templates, we train
IBAD by minimizing the Kullback--Leibler (KL) divergence between them, averaged over all
predictable target days:
\begin{equation}
   \mathcal{L}(\boldsymbol{\psi})
   = \frac{1}{|\mathcal{P}|}
     \sum_{(a_i, d) \in \mathcal{P}}
     \mathrm{KL}\!\bigl(\boldsymbol{\theta}^{a_i}_d \,\big\Vert\, \hat{\boldsymbol{\theta}}^{a_i}_d\bigr)
   = \frac{1}{|\mathcal{P}|}
     \sum_{(a_i, d) \in \mathcal{P}}
     \sum_{k=1}^{K} \theta^{a_i}_{d,k}\,
     \log \frac{\theta^{a_i}_{d,k}}{\hat{\theta}^{a_i}_{d,k}},
   \label{eq:ibad_loss}
\end{equation}
where $\boldsymbol{\psi}$ denotes all IBAD parameters and
\[
   \mathcal{P} = \bigl\{ (a_i, d) : d \ge 2,\; \boldsymbol{\theta}^{a_i}_d \neq \mathbf{0} \bigr\}
\]
restricts the sum to target days that have at least one day of history; padded days, marked
by $\boldsymbol{\theta}^{a_i}_d = \mathbf{0}$, are excluded. Equation~\eqref{eq:ibad_loss}
reduces to the standard categorical cross-entropy when $\boldsymbol{\theta}^{a_i}_d$ is
one-hot, so the soft-target formulation generalizes hard template prediction. The KL score is further used for anomaly scoring during inference.

\section{Experiments}
\label{sec:experiments}
This section evaluates whether modeling human mobility as mixtures of recurring
behavioral templates enables effective behavioral anomaly detection.
Specifically, we seek to answer three questions:

\begin{itemize}[leftmargin=*]
    \item \textbf{RQ1:} Does daily human mobility exhibit interpretable and recurring behavioral templates?
    \item \textbf{RQ2:} Does IBAD compete against existing anomaly detection methods in consistently identifying behavioral anomalies?
    \item \textbf{RQ3:} How robust is IBAD across other datasets?
\end{itemize}
We first describe the datasets and benchmark construction process, including
the Geolife semantic-labeling pipeline and our splicing-based anomaly detection benchmark. We then evaluate the behavioral templates discovered by LDA and compare IBAD's anomaly detection performance against state-of-the-art baselines.

\subsection{Datasets}
\label{sec:datasets}
\subsubsection{NHTS}
\label{sec:NHTS_dataset}
The National Household Travel Survey (NHTS) 2017 is a nationwide U.S. travel
survey conducted by the Federal Highway Administration, with data collected from
April 2016 through April 2017 and comprising $923{,}572$ trip records~\cite{FHWA2017NHTS}.
Each participant reports all of their trips within a single $24$-hour day,
including origin and destination activity purposes, departure and arrival times
in \texttt{HH:MM} format, and dwell time at each destination. NHTS is
\emph{cross-sectional}: each individual is observed on exactly one travel day. We used this to investigate the feasibility of discovering behavioral templates in Section~\ref{subsec:empirical_observations_templates_NHTS}. Because individuals were recorded for only a single day, this data is \textbf{not applicable} for directly testing our anomaly detection.

\subsubsection{NUMOSIM}
\label{sec:NUMOSIM_dataset}
NUMOSIM is a large-scale synthetic mobility dataset modeling the movements of $200{,}000$
agents across Los Angeles County over an eight-week period beginning in January 2024~\cite{stanford2024numosim}. It
includes two types of injected anomalies: non-recurring disruptions, which capture
one-off deviations from an individual's regular locations, and recurring deviations, which
represent systematic shifts in the timing of visits to those locations. Although NUMOSIM
models anomalies along both the spatial and temporal dimensions, it does not capture
genuine behavioral shifts that realistically reflect real-world applications of anomaly detection in human mobility.

\subsubsection{Geolife}
\label{sec:Geolife_dataset}
Geolife is a real-world GPS trajectory dataset collected by Microsoft Research Asia, primarily in Beijing, between 2007 and 2012~\cite{zheng2011geolife}. Unlike NHTS, which captures only a single travel day per participant, or NUMOSIM, which synthesizes mobility traces under controlled simulation conditions, Geolife provides long-term, real-world mobility traces for each individual spanning multiple years. This longitudinal structure makes it particularly valuable for modeling personal behavioral routines and detecting deviations from them over time. However, Geolife is inherently irregular, as daily observations are often missing.

\begin{table*}[t]
\centering
\caption{Dataset statistics. Agent-days counts are after preprocessing.
  Splice counts indicate the number of successfully injected anomalous sequences
  per policy: \emph{Intra},
  \textit{Inter}, and \emph{combined}.}
\label{tab:datasets}
\setlength{\tabcolsep}{5pt}
\begin{tabular}{lllrrrrrrr}
\toprule
\multirow{2}{*}{\textbf{Dataset}} &
\multirow{2}{*}{\textbf{Type}} &
\multirow{2}{*}{\textbf{Region}} &
\multirow{2}{*}{\textbf{Agents}} &
\textbf{Training} &
\textbf{Test} &
\textbf{Activity} &
\multicolumn{3}{c}{\textbf{\# Splice Anomalies (Agent Days)}} \\
\cmidrule(lr){8-10}
& & & & \textbf{Agent-Days} & \textbf{Agent-Days} & \textbf{Categories} &
\textbf{Intra} & \textbf{Inter} & \textbf{Combined} \\
\midrule
NHTS 2017  & Real      & Entire USA        & 264,234 & 212,734$^\dagger$ & N/A       & 18 & N/A   & N/A    & N/A    \\
NUMOSIM    & Synthetic & Los Angeles, CA   & 200,000 & 5,785,961         & 5,600,000 & 28 & 200,000 & 200,000 & 200,000 \\
Geolife    & Real      & Beijing, China    & 70      & 5,700             & 790       & 17 & 69     & 70      & 70      \\
\bottomrule
\end{tabular}
\vspace{1ex} 
\raggedright \footnotesize $^\dagger$ NHTS is cross-sectional, and each agent provides exactly one day of data. Therefore, training person-days entails the entire data used for template discovery.
\end{table*}

\subsection{Benchmark Construction}
\subsubsection{Geolife Semantic Activity Benchmark}
\label{sec:geolif_semantic_poi}
Geolife provides long-term GPS trajectories but lacks semantic activity labels,
making it unsuitable for behavioral anomaly detection in its raw form. To enable
behavioral analysis, we construct a semantic activity benchmark by converting
GPS traces into stay-points and assigning each stay one of the 18 NHTS activity
categories using geographic context from OpenStreetMap and LLM-based activity
inference. The resulting benchmark contains $70$ agents, $18{,}739$ stay-points, and 6,490 agent-days. This benchmark allows behavioral routines to be analyzed in terms of activity semantics rather than geographic coordinates alone. Complete details of the benchmark construction pipeline are provided in
Appendix~\ref{sec:geolife_POI_attribution}.

\subsubsection{Splicing Benchmark for Anomaly Detection} 
Since genuine daily behavioral anomaly labels are unavailable for Geolife and NUMOSIM, we construct a controlled benchmark via trajectory splicing. Specifically, we generate synthetic anomalies by replacing a single day in an agent’s trajectory with a day sampled from either the same agent or a different agent. This procedure induces controlled deviations in the agent’s behavioral template distribution, which is the core representation used by IBAD.

\noindent We define three splicing regimes to capture different types of behavioral shifts. \emph{Intra-agent splicing} replaces a day with another day from the same individual but from a different temporal context (e.g., weekday versus weekend), modeling within-person deviations from habitual routines. \emph{Inter-agent splicing} replaces a day with one drawn from a different individual, introducing cross-person behavioral mismatches. Finally, \emph{Combined splicing} randomly mixes both strategies to produce a heterogeneous setting.

\noindent To ensure that the injected days represent meaningful deviations in behavioral space rather than near-duplicate routines, we filter candidate substitutions using a Jensen--Shannon divergence constraint between behavioral template (topic-mixture) distributions. A substitution is accepted only if the divergence exceeds a small threshold $\varepsilon$, ensuring that each injected day represents a substantive shift in behavioral structure. Each splicing operation modifies exactly one day per agent. We assign binary labels at the day level ($1$ for injected anomalies and $0$ for normal days), yielding a standard daily behavioral anomaly detection task.


Appendix~\ref{sec:exp-setup} details the experimental setup on benchmarks.

\subsection{Baselines}
\label{sec:baselines}
We compare IBAD against conventional and
state-of-the-art anomaly detection methods.  
\begin{itemize}[leftmargin=*]
    \item \textbf{OCSVM}~\cite{scholkopf1999support} (One-Class Support Vector Machine) is an unsupervised learning algorithm that detects anomalies by modeling the support of a normal data distribution. It maps input features into a high-dimensional space using a Radial Basis Function (RBF) kernel to construct a tight decision boundary that encapsulates the normal training samples. Consequently, any observations falling outside this established frontier are classified as anomalous.

    \item \textbf{IF}~\cite{liu2008isolation} (Isolation Forest) is an ensemble-based anomaly detection algorithm that explicitly isolates anomalies rather than profiling normal data points. It recursively partitions the feature space using random splits, identifying anomalies as instances that require significantly shorter path lengths to be isolated within the resulting tree structures.
    
    \item \textbf{GaussianHMM}~\cite{rabiner1990tutorial}  represents each day as a sequence of $96$ temporal slot observations generated by a latent activity-state process. The model is trained on normal trajectories using the Baum--Welch algorithm~\cite{baum1970maximization}, and anomaly scores are computed from the negative log-likelihood of observed daily sequences.
    
    \item \textbf{DeepSVDD}~\cite{pmlr-v80-ruff18a} learns a compact representation of normal behavior by training a neural encoder to map normal samples close to a predefined latent center. Anomaly scores are computed as the squared distance from this center in latent space. We employ a lightweight MLP encoder over flattened daily representations.
    
    \item \textbf{TOD4Traj}~\cite{zhang2024transferable} employs feature-level and trajectory-level contrastive learning objectives to fuse spatial, temporal, and semantic information, capturing repetitive mobility patterns within and across agents. The model yields two complementary anomaly scores quantifying, respectively, an individual's deviation from its own regular visit patterns and its deviation from normal behavior at the population level.
    
    
    \item \textbf{BeSTAD}~\cite{xie2025bestad} learns personalized embeddings of recurring spatiotemporal mobility patterns for each individual and detects anomalies as deviations from these user-specific behavioral profiles. Consequently, the same visit pattern may be considered normal for one individual but anomalous for another.
    
    \item \textbf{ICAD}~\cite{azarijoo2025icad} is a multi-task autoregressive transformer that jointly predicts the next visit's activity, travel time, and departure time using categorical and GMM distributions in discrete activity categories and continuous time, respectively. 
    
    \item \textbf{TrajGenAgent}~\cite{li2026trajgenagent} is an LLM-based agentic framework originally designed for zero-shot trajectory generation, which we adapt for anomaly detection by computing the normalized edit distance between an agent's observed activity chain and a model-generated normal daily routine. 

\end{itemize}

\subsection{RQ1: Behavioral Template Discovery}
\label{sec:template_discovery_exp}
\begin{figure*}[ht]
    \centering
    \begin{subfigure}{\linewidth}
        \centering
        \includegraphics[width=\linewidth]{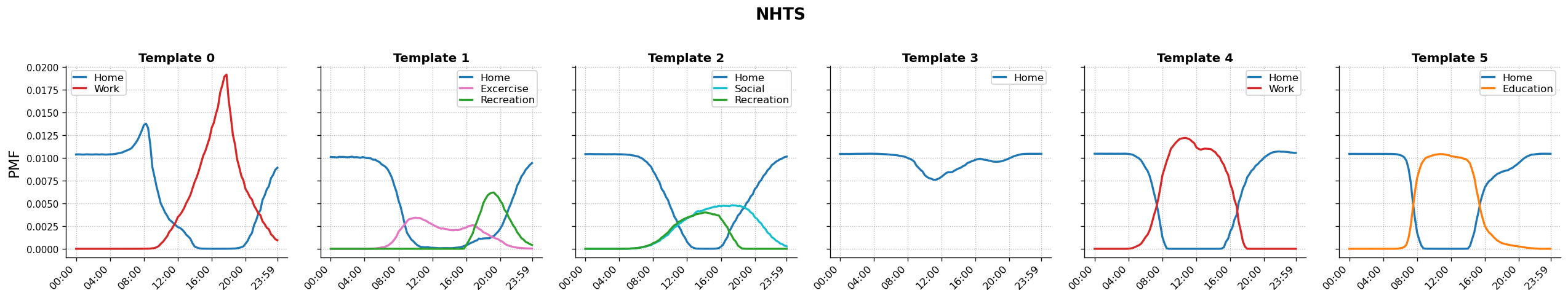}
        \caption{NHTS Behavioral Templates}
        \label{fig:nhts_templates}
    \end{subfigure}

    \vspace{0.5em}

    \begin{subfigure}{\linewidth}
        \centering
        \includegraphics[width=\linewidth]{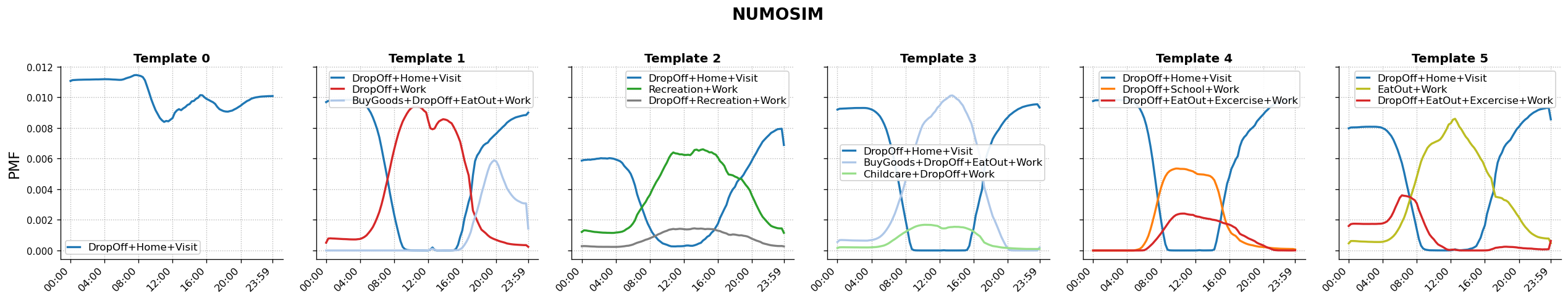}
        \caption{NUMOSIM behavioral templates.}
        \label{fig:numosim_templates}
    \end{subfigure}

    \vspace{0.5em}

    \begin{subfigure}{\linewidth}
        \centering
        \includegraphics[width=\linewidth]{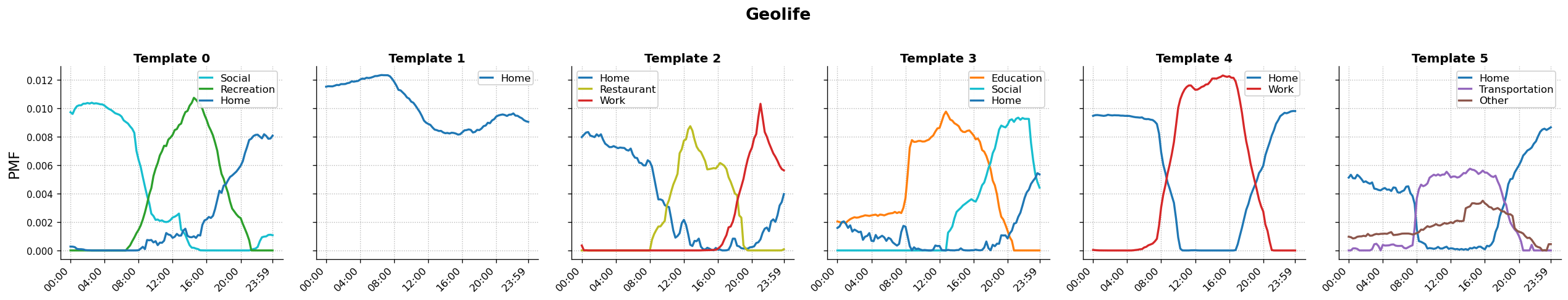}
        \caption{Geolife Behavioral Templates.}
        \label{fig:geolife_templates}
    \end{subfigure}

    \caption{Visualization of latent behavioral templates extracted using LDA across the three datasets. Each template represents a recurring temporal activity pattern learned from daily mobility trajectories.}
    \label{fig:lda_templates}
\end{figure*}

In this section, we examine whether daily human mobility data naturally decompose into
interpretable and recurring behavioral templates, which is the central
assumption underlying IBAD. To this end, we apply LDA to time-slotted activity
sequences and represent each day as a mixture of latent behavioral templates
rather than a single activity category. 

Figure~\ref{fig:lda_templates} visualizes the extracted behavioral templates for each dataset. For clarity, we display only the dominant activities associated with each template over a 24-hour period. The $x$-axis represents the time of day, while the $y$-axis denotes the joint probability of observing a specific activity at a particular time conditioned on the corresponding template. These temporal probability distributions reveal semantically meaningful behavioral structures. For instance, certain templates exhibit concentrated daytime work-related activity corresponding to conventional work schedules, whereas others capture evening recreational behavior, commuting periods, or home-centered routines. In summary, we observe:
\begin{itemize}[leftmargin=*]
    \item Several behavioral patterns are common across datasets. Most
    templates begin and end with \texttt{Home} activity, reflecting overnight residence.
    Likewise, a work-centered template emerges in all datasets (NHTS
    Template 4, NUMOSIM Template 1, and Geolife Template 4), characterized by \texttt{Work} activity
    during regular business hours. The recurrence of this pattern across
    travel survey, NUMOSIM, and Geolife data suggests that work routines constitute a
    stable and transferable behavioral archetype. In addition, NHTS and Geolife templates can discriminate  part-time shift workers from regular full-time workers, as shown in Template 0 for NHTS and Template 2 for Geolife.
    \item  The templates reveal a strong temporal organization of daily behavior.
    Among datasets, discretionary activities such as \texttt{Recreation}, \texttt{Social},
    \texttt{Restaurant}, and \texttt{Exercise} are concentrated in the afternoon and evening
    (approximately 12:00--22:00 PM), whereas mornings are dominated by \texttt{Home}
    and transitions associated with work or school. This separation suggests that
    the discovered templates capture meaningful behavioral semantics rather than
    simple activity frequencies. In particular, the concentration of discretionary
    activities into distinct templates indicates that human mobility is not
    arbitrarily distributed throughout the day but instead follows predictable
    temporal structures. Such regularity is precisely what enables deviations from
    routine behavior to be identified as anomalies.
    \item  Because NUMOSIM labels POIs with aggregated activity bundles (e.g., \texttt{DropOff+School+Work}) rather than assigning visit-specific activity purposes to individual agents, the resulting templates are inherently less interpretable. This location-level aggregation makes it impossible to differentiate underlying behavioral intents at the same venue, such as distinguishing a teacher arriving for employment (\texttt{Work}) from a student arriving for instruction (\texttt{Education}). In contrast, NHTS and Geolife yield cleaner
    single-purpose templates. Geolife additionally contains a
    Transportation-dominated template that is absent from NHTS and NUMOSIM,
    reflecting the fact that GPS trajectories explicitly capture in-transit behavior.
\end{itemize}

Overall, the discovered templates are both interpretable and consistent across
datasets, supporting the hypothesis that daily mobility lies within a space of a few behavioral archetypes. This finding validates the core premise of IBAD: normal behavior can be represented as mixtures of a small number of behavioral templates, while anomalies correspond to departures from these learned templates. 

\subsection{RQ2: Anomaly Detection Results}
\begin{table*}[t]
\centering
\small
\setlength{\tabcolsep}{6pt}
\caption{Splicing benchmark results across datasets and policies. AUROC and AP are reported separately. Bold indicates the best performance and underline indicates the second-best performance within each policy setting.}
\begin{tabular}{lcccccccccccc}
\toprule
\multirow{3}{*}{\textbf{Model}}
& \multicolumn{6}{c}{\textbf{NUMOSIM}}
& \multicolumn{6}{c}{\textbf{GeoLife}} \\
\cmidrule(lr){2-7}
\cmidrule(lr){8-13}
& \multicolumn{2}{c}{\textbf{Intra}}
& \multicolumn{2}{c}{\textbf{Inter}}
& \multicolumn{2}{c}{\textbf{Combined}}
& \multicolumn{2}{c}{\textbf{Intra}}
& \multicolumn{2}{c}{\textbf{Inter}}
& \multicolumn{2}{c}{\textbf{Combined}} \\
\cmidrule(lr){2-3}
\cmidrule(lr){4-5}
\cmidrule(lr){6-7}
\cmidrule(lr){8-9}
\cmidrule(lr){10-11}
\cmidrule(lr){12-13}
& AUROC & AP
& AUROC & AP
& AUROC & AP
& AUROC & AP
& AUROC & AP
& AUROC & AP \\
\midrule
Isolation Forest
& \textbf{0.7991} & \underline{0.0651}
& \underline{0.8867} & 0.1383
& \underline{0.8835} & 0.1306
& 0.5847 & 0.0459
& 0.6149 & 0.0747
& 0.6279 & 0.0726 \\
GaussianHMM
& 0.7326 & 0.0467
& 0.8638 & \underline{0.1421}
& 0.8633 & \underline{0.1316}
& 0.5753 & 0.0494
& 0.5972 & 0.0872
& 0.6274 & 0.0725 \\
OCSVM
& 0.7533 & \textbf{0.0699}
& 0.8429 & 0.1143
& 0.8400 & 0.1079
& 0.5305 & 0.0374
& 0.5354 & 0.0649
& 0.5052 & 0.0617 \\
DeepSVDD
& 0.5989 & 0.0477
& 0.6416 & 0.0712
& 0.6555 & 0.0729
& 0.5670 & 0.0442
& 0.5906 & 0.0689
& 0.5749 & 0.0718 \\
TOD4Traj
& 0.5051 & 0.0147
& 0.5673 & 0.0212
& 0.5651 & 0.0211
& \textbf{0.6533} &\underline{0.0643}
& \textbf{0.7109} & 0.1018
& \textbf{0.7336} & 0.0936 \\
BeSTAD
& 0.5043 & 0.0156
& 0.5003 & 0.0188
& 0.5002 & 0.0188
& 0.5308 & 0.0443
& 0.6297 & \textbf{0.1562}
& 0.6128 & \textbf{0.1423} \\
ICAD
& 0.6033 & 0.0217
& 0.7756 & 0.1064
& 0.7342 & 0.0773
& 0.4842 & 0.0336
& 0.4813 & 0.0490
& 0.5657 & 0.0550 \\

TrajGenAgent
& 0.5994 & 0.0241
& 0.7730 & 0.0688
& 0.7496 & 0.0593
& 0.5903 & 0.0461
& 0.5026 & 0.0500
& 0.5751 & 0.0568 \\

IBAD
& \underline{0.7918} & 0.0552
& \textbf{0.8904} & \textbf{0.1471}
& \textbf{0.8875} & \textbf{0.1399}
& \underline{0.5974} & \textbf{0.0725}
& \underline{0.7079} & \underline{0.1213}
& \underline{0.6404} & \underline{0.0938} \\
\bottomrule
\end{tabular}
\label{tab:splice_benchmark}
\end{table*}

Having discovered interpretable behavioral archetypes, in this section, we present anomaly detection results of IBAD compared to baselines. Table~\ref{tab:splice_benchmark} summarizes results on the NUMOSIM and Geolife datasets across all splicing policies. Overall, IBAD achieves either the best or second-best performance in 11 of the 12 settings, demonstrating robustness across datasets and anomaly types. In summary:

\begin{enumerate}[leftmargin=*,itemsep=2pt,topsep=2pt]
\item \noindent \textbf{Datasets Analysis.} IBAD achieves substantially higher scores in NUMOSIM compared to Geolife, particularly in AUROC. For instance, IBAD attains AUROC scores of 0.79–0.89 on NUMOSIM, but only 0.59–0.71 on Geolife. This trend is visible across other baselines, indicating Geolife is a more challenging dataset. One possible explanation is extreme sparsity of Geolife, which has only 70 agents and many missing days in an agent's temporal window. Such sparsity restricts the availability of representative behavioral patterns and make anomaly detection much more difficult. Nevertheless, IBAD remains competetive on Geolife, achieving the highest AP under the \emph{Intra}-policy and second-best scores elsewhere.
\item \noindent \textbf{Policy Analysis.} Across both datasets and all methods, \emph{Inter}-splicing policy consistently holds the highest AUROC and AP scores, whereas \emph{Intra}-splicing policy has the lowest scores. This suggests that behavioral mismatches unexplained by a person's regular template mixture are straightforward to detect than anomalies created by splicing an individual's weekday and weekend routines. Since Intra-splicing preserves the agent's underlying behavioral preferences and mobility patterns, the spliced trajectories often resemble routine variations rather than behavioral anomalies. In contrast, Inter-splicing introduces behaviors that are inconsistent with the agent's behavioral profile, producing stronger deviations from expected routines. The \emph{Combined} policy generally exhibits intermediate performance, reflecting the mixture of both anomaly types.
\item \noindent \textbf{Baseline Analysis.} Classical anomaly detection methods such as IF and OCSVM perform relatively well on NUMOSIM, but their performance degrades on Geolife, indicating weaker generalization to sparse and irregular mobility data. GaussianHMM shows provide more stable performance across policies but still fail to capture complex individual behavioral shifts, resulting in moderate AUROC gains and relatively low AP scores, which is a sign of high false positives. Deep learning baselines show mixed behaviors across different policies and datasets. Notably, TOD4Traj perform competitively on Geolife but struggle on NUMOSIM, suggesting sensitivity to dataset structure and limited adaptability across domains. In addition, ICAD also fails on Geolife due to its original visit-level anomaly scoring, which depends on dense visit sequences and thus, fails to learn normal behavior effectively where visit recordings are sparse and irregularly spaced. In contrast, IBAD retains consistent scores in all settings. The reason is that IBAD combines agent-level behavioral template prediction module with a daily representation [CLS] vector that is invariant to visit density. Moreover, it conditions its template mixture on the target day, which is essentially useful in irregular datasets as it allows the model to learn time gaps effectively, and thus, enables generalizability across sparse and irregular datasets.
\end{enumerate}

\subsection{RQ3: Robustness under Distribution Shift}
\label{sec:transferability}
A major challenge in human mobility modeling is ensuring that learned behavioral archetypes generalize across geographic and demographic contexts rather than capturing dataset-specific characteristics such as spatial topology or population dynamics. By extracting semantic behavioral templates with LDA, we hypothesize that these templates capture universal human behavioral patterns rather than localized artifacts. To validate this, we show that \emph{cross-sectional} NHTS-discovered behavioral templates remain useful for anomaly detection on the real-world, \emph{longitudinal} Geolife dataset, whose distribution differs substantially from that of NHTS. We do not evaluate transferability of NHTS on NUMOSIM dataset as NUMOSIM lacks primary activity categories.  

\noindent We confirm cross-dataset transferability statistically by aligning the NHTS and Geolife templates using Hungarian matching~\cite{kuhn1955hungarian} and assessing the alignment against a permutation null, through which we find the cross-dataset template similarity is significantly above chance (see Appendix~\ref{app:transferability} for the full experiment). Table~\ref{tab:transferability} reports anomaly detection results of the transferred templates. Overall, \emph{Inter}-splicing yields the highest AUROC and its AP is comparable to Geolife is comparable to that of Geolife templates trained and evaluated on their own LDA fit, indicating that NHTS cross-sectional reference
templates are most effective at flagging \emph{semantic} mismatches in which an agent's overall daily behavior is inconsistent with its established routines. In contrast \emph{Intra}-splicing is the weakest among the three, with an AUROC of $0.465$, close to chance. This is expected as NHTS templates are derived from a single 24\,h snapshot of each person's mobility rather than from \emph{longitudinal} recordings of an individual's trips over an extended window, so they capture one day archetypes. Finally, the \emph{Combined} policy is intermediate in ranking quality AUROC of $0.544$, yet attains the highest AP of $0.114$, suggesting that combining both policies gives the most favorable precision–recall trade-off even though its AUROC ranking is not the sharpest.

\begin{table}[ht]
\centering\small
\caption{Reusability of NHTS behavioral
templates: daily anomaly detection on Geolife using NHTS soft templates. IBAD is trained on Geolife with NHTS templates.}\label{tab:transferability}
\setlength{\tabcolsep}{4pt}
\begin{tabular}{lcccccc}
\toprule
& \multicolumn{2}{c}{Inter} & \multicolumn{2}{c}{Intra} & \multicolumn{2}{c}{Combined} \\
\cmidrule(lr){2-3}\cmidrule(lr){4-5}\cmidrule(lr){6-7}
Transfer & AUROC & AP & AUROC & AP & AUROC & AP \\
\midrule
NHTS\,$\rightarrow$\,Geolife & 0.627 & 0.1038 & 0.465 & 0.0486 & 0.544 & 0.114 \\
\bottomrule
\end{tabular}
\end{table}

\section{Conclusion}
\label{sec:conclusion}
This work introduces a new perspective on mobility anomaly detection: natural spatiotemporal variability is not an anomaly, but rather expected diversity within a person's regular behavioral templates. Building on this intuition, IBAD utilizes \emph{topic modeling} to discover global behavioral templates and a hierarchical self-supervised model to learn an individual's expected template mixtures. During inference, it identifies anomalies by measuring deviations from these established routines. Experiments across real-world and synthetic datasets demonstrate IBAD's competitive performance. Ultimately, IBAD achieves competitive results in completely unlabeled data without relying on handcrafted dimensions, grounding the framework in realistic mobility dynamics. Future directions include extending this day-level architecture to support real-time inference. By transitioning from daily boundaries to sliding temporal windows, IBAD can dynamically evaluate partial-day template mixtures, enabling continuous behavioral anomaly detection.

\bibliographystyle{ACM-Reference-Format}
\bibliography{references}

\appendix

\section{Geolife Semantic POI Processing Pipeline}
\label{sec:geolife_POI_attribution}
\subsection{Stay-point Extraction} We construct a labeled mobility benchmark from the Microsoft Geolife GPS Trajectories v1.3 dataset, restricting to the greater Beijing bounding box (39.75°N–40.08°N, 116.06°E–116.72°E) to concentrate on the densely-sampled urban cohort following~\cite{zhang2024large}. Agents left with fewer than 100 position fixes after this spatial filter
are discarded to exclude sparse, unreliable traces. The retained raw GPS position fixes are processed by the sliding-window stay-point detection algorithm of ~\cite{li2008mining}, implemented by the \texttt{trackintel} library~\cite{Martin_2023_trackintel} with a spatial threshold of 200\,m, a minimum dwell time of 30\,min, and a gap-splitter threshold of 12\,h. The 12\,h gap threshold is deliberately high so that GPS-silence periods shorter than half a day (e.g. overnight stay at home) do not
fragment a single continuous visit into multiple records. After extraction, agents with fewer than 50 stay-points are removed, yielding \textbf{70 agents} and \textbf{18,739 stay-points}.

\subsection{POI Attribution} Behavioral anomaly detection requires semantic activity labels beyond raw coordinates: two agents visiting the same city block at the same hour may
look identical spatially, yet one is a worker at their regular workplace while the
other is a student and thus, follows a student's behavioral patterns. Therefore, we assign each stay-point a category from the 18-class NHTS 2017 activity taxonomy (Home, Work, Education, Childcare, Adultcare, Shopping, Services, Errands, Recreation, Restaurant, Religious, Social, Exercise, Health, Pickup\_Dropoff, Transportation, Volunteer,
Other) using a two-stage pipeline of OSM enrichment followed by an \emph{LLM-as-a-judge}~\cite{zheng2023judging,gilardi2023chatgpt}.

Each unique stay centroid, rounded to four decimal places ($\approx$11\,m at Beijing latitude), is first reverse-geocoded via the OpenStreetMap Nominatim API (\texttt{zoom=18}, \texttt{extratags=1})~\cite{olbricht2015data} to retrieve the dominant physical-feature tuple (\texttt{category}, \texttt{type}, \texttt{class}, \texttt{extratags}). Since OSM annotates the physical feature beneath a GPS fix rather than the activity performed there (e.g. a centroid snapped to a residential road conveys no useful behavioral signal), we additionally query the Overpass API~\cite{olbricht2015data} for the top-3 nearest activity-relevant POIs within 100\,m of each centroid, expanding to 500\,m in spatially sparse regions. Both OSM views are cached in a local SQLite database so each unique centroid is queried exactly once.

Each stay-point in a day of an agent is labeled in a single call to \texttt{Gemini~2.5~Flash}
acting as an \emph{LLM-as-a-judge}~\cite{gemini25_2025}. The model is provided with the NHTS activity taxonomy, the OSM-derived physical attributes and top-3 neighbor POIs for each stay, inter-stay transit distances and gap times, and a per-agent behavioral profile summarizing the five most-visited centroids with a temporal fingerprint: visit frequency, median arrival time with interquartile range, median dwell duration with IQR, weekday/weekend split, and calendar date span. To reduce ambiguity, the prompt enforces a six-step reasoning chain:
(\emph{1})~read the primary OSM tag;
(\emph{2})~apply duration and time-of-day heuristics
    (e.g., overnight stays at residential locations default to \texttt{Home};
    long regular daytime stays at non-residential sites default to \texttt{Work};
(\emph{3})~consult the top-3 nearest neighbor POIs as a secondary spatial signal;
(\emph{4})~check the agent's visit-history prior for that centroid;
(\emph{5})~disambiguate transit-like stays between
    \texttt{Transportation} and \texttt{Pickup\_Dropoff}; and
(\emph{6})~assign \texttt{Other} only when all preceding steps fail to
    yield a confident label. Because \texttt{Other} indicates insufficient evidence rather than a genuine residual category, we run a second labeling pass over every sample that contains at least one \texttt{Other} stay. In this pass, each frequent-location entry is augmented with its modal label from pass~1, providing the model with explicit within-agent consistency cues; the pass-1 label is retained only when pass~2 still returns \texttt{Other}. As shown in Table~\ref{tab:geolife_activity_dist}, 17 out of 18 NHTS activity types have stay-points assigned to it. 
        

\begin{table}[t]
\centering
\caption{Distribution of NHTS activity categories across the 18,739 Geolife stay-points (70 agents). Categories are ordered by frequency; \emph{Volunteer} is absent in Geolife POI activities.}
\label{tab:geolife_activity_dist}
\small
\setlength{\tabcolsep}{4pt}
\begin{tabular}{lrr@{\hspace{0.5em}}lrr}
\toprule
Activity & Count & \% & Activity & Count & \% \\
\midrule
Work           & 3,570 & 19.05 & Shopping       & 1,111 & 5.93 \\
Home           & 2,226 & 11.88 & Other          &   682 & 3.64 \\
Recreation     & 2,009 & 10.72 & Services       &   571 & 3.05 \\
Restaurant     & 1,765 &  9.42 & Pickup/Dropoff &   379 & 2.02 \\
Social         & 1,763 &  9.41 & Health         &   328 & 1.75 \\
Education      & 1,539 &  8.21 & Exercise       &   274 & 1.46 \\
Transportation & 1,232 &  6.57 & Childcare      &    80 & 0.43 \\
Errands        & 1,176 &  6.28 & Religious      &    30 & 0.16 \\
               &       &       & Adultcare      &     4 & 0.02 \\
\midrule
\multicolumn{6}{c}{Total: 18,739 stay-points across 17 shared categories} \\
\bottomrule
\end{tabular}
\end{table}

\section{Experimental Setup}
\label{sec:exp-setup}
\subsection{LDA Templates.}
\label{sec:lda_templates}
For every dataset we fit a \texttt{sklearn} LDA model with $K{=}6$ topics using the Online Variational solver while dividing daily hours in 15 minute time intervals, leading to 96 slots covering the entire day. The reason for selecting 15 minute time intervals is accommodating LDA to learn reach behavioral templates from short and long visits. It applies to the \emph{cross-sectional} NHTS 2017 travel survey, and \emph{longitudinal} Geolife and NUMOSIM datasets. The NHTS and Geolife vocabularies are slot–activity tokens ($96$ slots $\times$ $M$ activities). For NUMOSIM, since we do not have one unique activity category for each POI, we bundled all unique set of activity categories, resulting in 28  bundles, and treated each bundle as one activity category. We then Fit LDA on the entire NHTS dataset, training set of Geolife, and $200{,}000$ random daily agent sequences of NUMOSIM.

\subsection{Anomaly Detection Datasets.}
IBAD is trained and evaluated on \emph{longitudinal} NUMOSIM and Geolife datasets. For NUMOSIM, we follow the original benchmark protocol: a $28$-day training window followed by a $28$-day test window, with no validation split. Since NUMOSIM does not annotate the primary activity of each visit, we construct an activity vocabulary from the POIs themselves: at every location we collect the set of activities its POI affords, and treat each distinct set as a single bundle. This produces $28$ unique bundles, and every visit is labeled with the bundle of its POI. Geolife is split chronologically per agent into $80/10/10$ train/validation/test, while the entire cross-sectional NHTS~2017 is used for template discovery. The value of $\varepsilon$ was set to 0.005 for NUMOSIM and 0.05 for Geolife .Spliced anomalies are injected only into the test splits of NUMOSIM and Geolife, because the NHTS~2017 data has only one day of observation for each agent.

\subsection{Hardware Configuration}
\label{sec:hardware}
All experiments are run on a workstation with two AMD EPYC 7763 
$64$-core CPUs and eight NVIDIA RTX $6000$ Ada Generation GPUs ($48$\,GB GDDR6 each); a single GPU is used per training run. The  stack includes Python $3.12$,  PyTorch $2.9.1$ and CUDA $12.8$.

\subsection{Evaluation Metrics}
To evaluate IBAD's effectiveness in identifying unusual behavioral shifts from an agent's regular norm, we consider Average Precision (AP) and Area Under Receiver Operator Characteristic (AUROC). AP is computed as the weighted average of precisions across different recall levels and is suitable for measuring extremely rare anomalies in highly imbalanced datasets~\cite{liu2020online, zheng2017contextual, davis2006relationship}, whereas AUROC measures a model’s ability to assign higher scores to true anomalies regardless of class imbalance. Since these metrics are \emph{ranking}-based, they evaluate the relative ordering of anomaly scores rather than relying on a fixed decision threshold, making them well suited for unsupervised anomaly detection.


\subsection{IBAD}
\label{sec:IBAD_arch}
We use two model sizes, scaled to the dataset volume. The NUMOSIM configuration has embedding dimension $d_{\text{embed}}{=}32$, visit dimension $d_v{=}64$, model dimension $d_{\text{model}}{=}128$, $L{=}6$ day-level Transformer layers with $H{=}8$ attention heads, and $L_v{=}4$ visit-level layers with $H_v{=}4$ heads. For Geolife, we adopt a lighter configuration with $d_{\text{embed}}{=}16$, $d_v{=}32$, $d_{\text{model}}{=}64$, and $L{=}2$ day-level layers. All other architectural choices are shared across datasets:  time-decayed attention, attention-based day pooling, dropout  $= 0.2$, and a Space2Vec encoder for latitude–longitude inputs. The model is optimised with AdamW (weight decay $10^{-4}$) under a cosine-annealed learning-rate schedule. NUMOSIM runs for $100$ epochs at batch size $4096$ and learning rate $10^{-3}$. Geolife runs for up to $300$ epochs at batch size $32$ and learning rate $5{\times}10^{-4}$, with KL-based early stopping on the validation split (patience $50$). All runs use random seed $42$ for reproducibility.

\section{Behavioral Template Transferability}
\label{app:transferability}
\subsection{Cross-Dataset Template Alignment}
To assess whether the behavioral templates identified by the LDA model in Section~\ref{sec:template_extraction} represent genuine, dataset-independent archetypes, we compare templates learned independently from the NHTS and Geolife datasets. Specifically, we fit the template model separately to each dataset, obtaining two sets of templates,
$\{\boldsymbol{\phi}^{\text{NHTS}}_k\}_{k=1}^{K}$ and
$\{\boldsymbol{\phi}^{\text{Geo}}_k\}_{k=1}^{K}$, with $K=6$.
We then measure the agreement between the two template sets. Since NHTS and Geolife share 17 activity categories from original 18 activity categories of NHTS, we confine the comparison to the common categories while keeping the full set of $S$ time slots. Under this assumption, two templates are considered similar only if they assign similar probabilities to the \emph{same activities at the same times of day}. Also, it is important to mention template indices are arbitrary across two independent fits, so
we first recover the best one-to-one correspondence between the two sets with Hungarian matching~\cite{kuhn1955hungarian} on their pairwise cosine similarities, and report the mean cosine of the matched pairs, which we denote by $\bar{s}=0.65$. Furthermore, Figure~\ref{fig:template_transferability_pairs} illustrates per-template cosine similarity of matched Geolife-NHTS templates of Section~\ref{sec:template_discovery}. For instance, $T3$ in NHTS and $T_1$ in Geolife both correspond to dominant \emph{stay-at-home} patterns, $T_4$ in both datasets characterize a \emph{full-time worker} behavioral profile, and $T_5$ in NHTS and $T_3$ in Geolife characterize a student behavior. These high similarities signal universal transferability of across nations. To prove this, we further conduct a statistical significance test in \ref{sec:hypotheis_tranferability_test}.

\subsection{Statistical Significance of Template Alignment}
\label{sec:hypotheis_tranferability_test}
While the mean matched similarity $\bar{s}$ summarizes the agreement between the NHTS and Geolife template sets, its magnitude alone is not directly interpretable. Some overlap is expected by chance because activities such as \texttt{Home} account for a large share of activity--time slots in both datasets. Consequently, elevated similarity scores may arise even in the absence of shared behavioral structure. To assess whether the observed alignment reflects genuine, transferable mobility patterns, we compare it against a permutation-based null model.

\begin{figure}[htbp]
\centering

\begin{subfigure}{\linewidth}
    \centering
    \includegraphics[width=\linewidth]{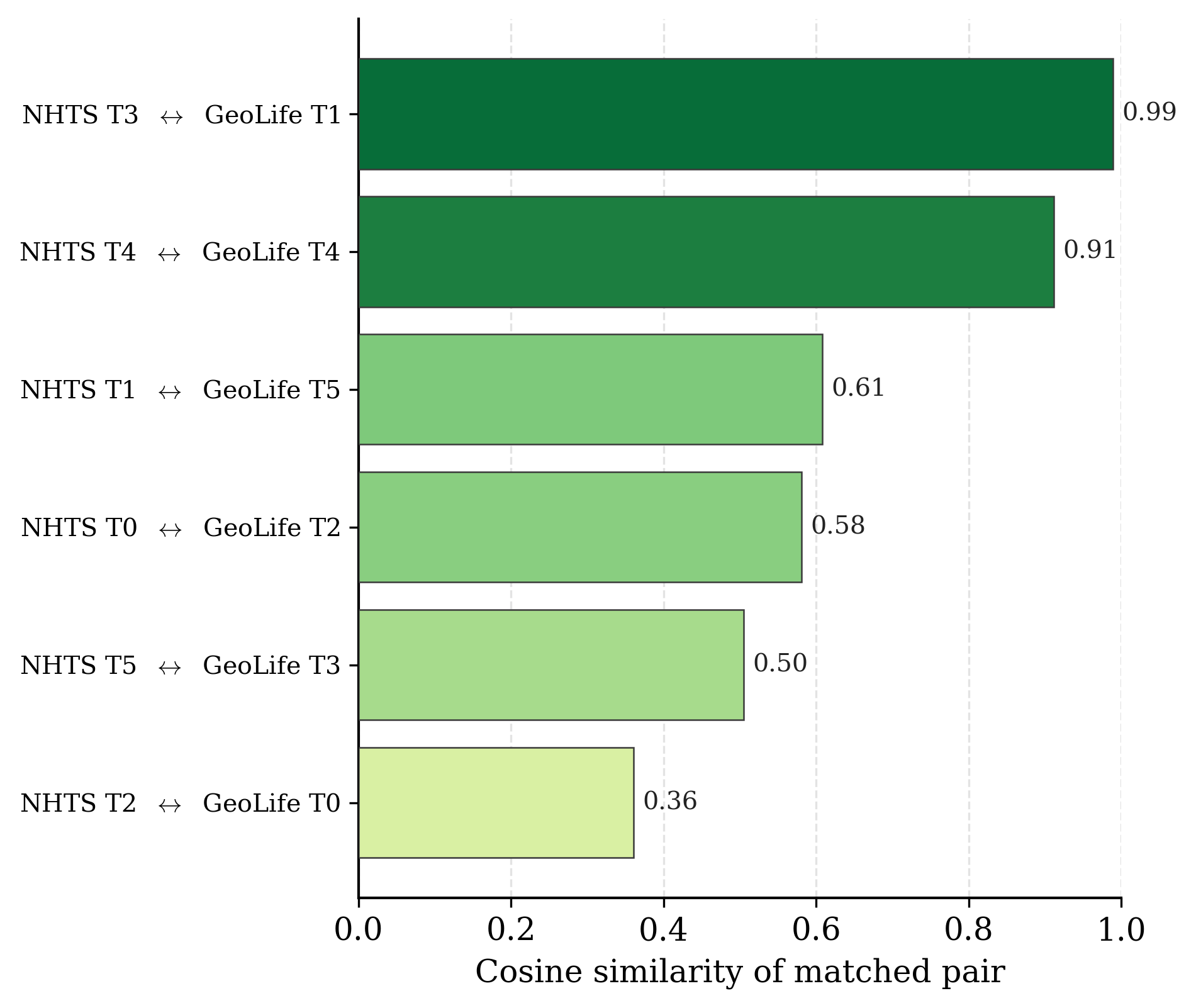}
    \caption{Cosine similarity of each matched NHTS--Geolife template pair after Hungarian matching. $T_i$ denotes the template identifiers from Figure~\ref{fig:lda_templates}.}
    \label{fig:template_transferability_pairs}
\end{subfigure}
\vspace{0.5em}
\begin{subfigure}{\linewidth}
    \centering
    \includegraphics[width=\linewidth]{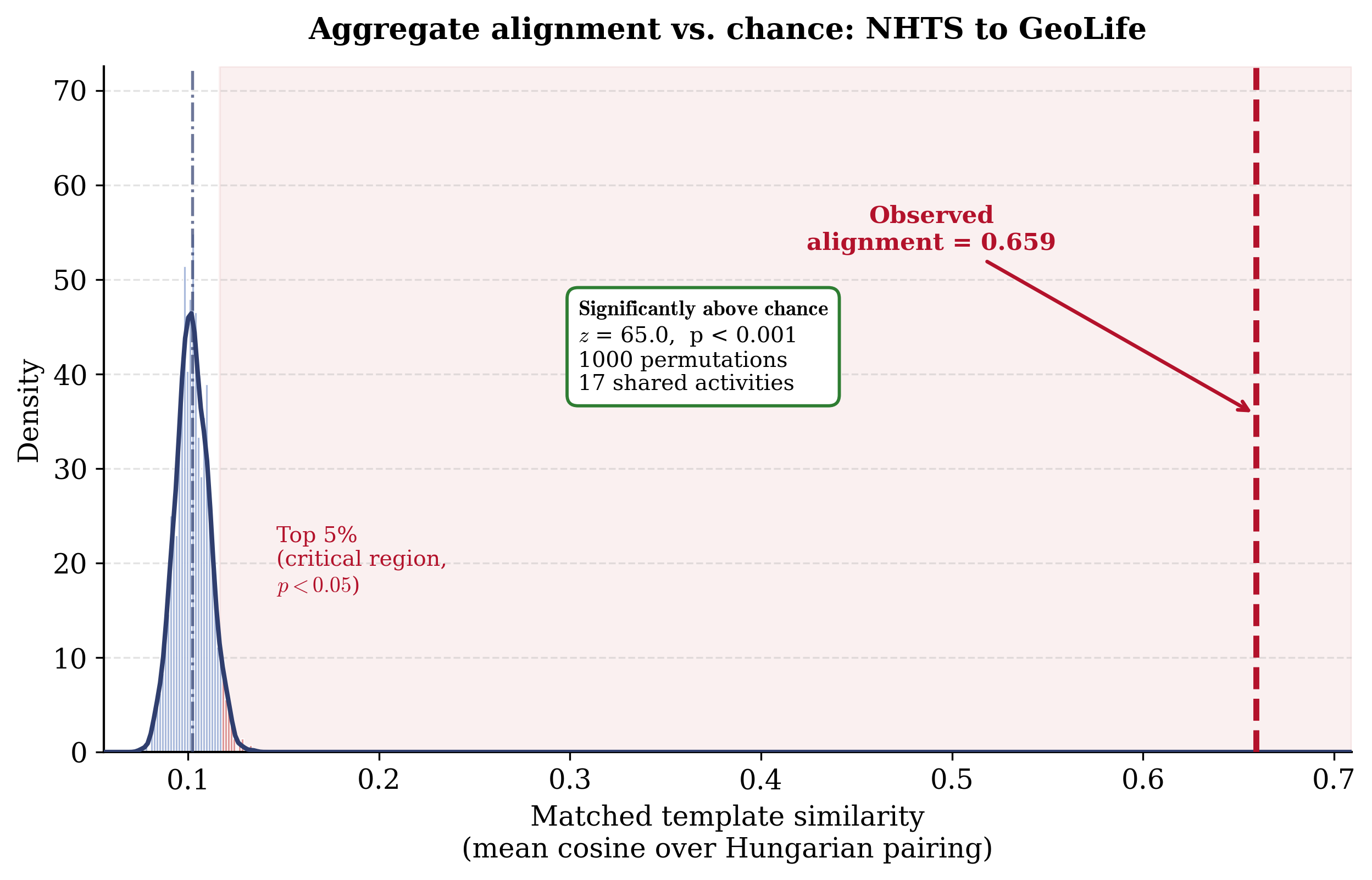}
    \caption{Permutation-based null distribution of the mean matched cosine similarity. The vertical red line indicates the observed alignment score.}
    \label{fig:template_transferability_distribution}
\end{subfigure}

\caption{Cross-dataset template alignment between NHTS and Geolife. (b) Similarity of individual matched template pairs after optimal matching. (b) Comparison of the observed overall matched similarity against a null distribution. }
\label{fig:template_transferability}
\end{figure}

To construct this null distribution, we generate 1,000 randomized versions of the Geolife templates. For each template, the $(s,a)$ entries are randomly permuted, preserving the overall distribution of activity probabilities while eliminating the temporal structure linking activities to specific times of day. After each randomization, we repeat the Hungarian matching procedure and compute the resulting mean cosine similarity between matched template pairs. This yields the level of alignment expected when activity--time associations are absent. Figure~\ref{fig:template_transferability_distribution} shows the resulting null distribution together with initial observed alignment score of $\bar{s}=0.659$, is substantially larger than the similarities obtained under randomization, which are concentrated around $0.10$. None of the 1,000 permutations produced a similarity as large as the observed score, corresponding to $p<0.001$. To quantify the magnitude of this departure from the null distribution, we compute a standardized effect size:
\[
z = \frac{\bar{s} - \mu_{\text{null}}}{\sigma_{\text{null}}},
\]
where $\mu_{\text{null}}$ and $\sigma_{\text{null}}$ denote the mean and standard deviation of the permutation distribution. Intuitively, $z$ measures how many standard deviations the observed alignment lies above the level expected by chance. In our case, $z \approx 65$, indicating that the observed similarity is extraordinarily far from the null expectation.

Together, these results provide strong evidence that the discovered templates represent true transferable daily activity archetypes rather than dataset-specific artifacts. The significant alignment between independently learned template sets supports the use of these templates across datasets in the transferability analysis of Section~\ref{sec:transferability}.

\clearpage
\end{document}